\documentclass[sigconf]{acmart}

\renewcommand\footnotetextcopyrightpermission[1]{} 
\usepackage{booktabs} 

\setcopyright{rightsretained}

\usepackage{graphicx}
\usepackage{amsmath,amssymb,amsfonts,amsthm}
\usepackage{mathtools}
\usepackage[ruled,linesnumbered]{algorithm2e}
\usepackage{url}

\acmArticle{4}
\acmPrice{15.00}

\begin{document}
\title{Future Automation Engineering using Structural Graph Convolutional Neural Networks}

\author{Jiang Wan}
\authornote{This work was performed while these authors were at Siemens.}
\affiliation{
  \institution{University of California Irvine}
}
\email{jiangwan@uci.edu}

\author{Blake S.\ Pollard${^*}$}
\affiliation{
  \institution{Carnegie Mellon University}
}
\email{blake561@gmail.com}

\author{Sujit Rokka Chhetri${^*}$}
\affiliation{
  \institution{University of California Irvine}
  }
\email{schhetri@uci.edu}

\author{Palash Goyal${^*}$}
\affiliation{
  \institution{University of Southern California}
  }
\email{palashgo@usc.edu}

\author{Mohammad Abdullah Al Faruque}
\affiliation{
  \institution{University of California Irvine}
 }
\email{alfaruqu@uci.edu}

\author{Arquimedes Canedo}
\affiliation{
 \institution{Siemens Corporate Technology}
 }
\email{arquimedes.canedo@siemens.com}

\begin{abstract}
The digitalization of automation engineering generates large quantities of engineering data that is interlinked in knowledge graphs. Classifying and clustering subgraphs according to their functionality is useful to discover functionally equivalent engineering artifacts that exhibit different graph structures. This paper presents a new graph learning algorithm designed to classify engineering data artifacts -- represented in the form of graphs -- according to their structure and neighborhood features. Our Structural Graph Convolutional Neural Network (SGCNN) is capable of learning graphs and subgraphs with a novel graph invariant convolution kernel and downsampling/pooling algorithm. On a realistic engineering-related dataset, we show that SGCNN is capable of achieving $\approx$91\% classification accuracy.
\end{abstract}

%
%


\copyrightyear{2018} 
\acmYear{2018} 
\setcopyright{acmcopyright}
\acmConference[ICCAD '18]{IEEE/ACM INTERNATIONAL CONFERENCE ON COMPUTER-AIDED DESIGN}{November 5--8, 2018}{San Diego, CA, USA}
\acmBooktitle{IEEE/ACM INTERNATIONAL CONFERENCE ON COMPUTER-AIDED DESIGN (ICCAD '18), November 5--8, 2018, San Diego, CA, USA}
\acmPrice{15.00}
\acmDOI{10.1145/3240765.3243477}
\acmISBN{978-1-4503-5950-4/18/11}

\maketitle

\section{Introduction}
The engineering of a complex system is a lengthy and complex process that generates a very large, continuous, and asymmetric influx of data. Automation engineering refers to the design, creation, development and management of production systems in factories, process plants, and supply chains that realize the production of products. The \textit{future of automation engineering} relies on the digitalization of the engineering process such that all engineering data from the product and its production system is captured and interlinked. Based on the recent successes of knowledge graphs for knowledge representation in search~\cite{KnowledgeVault,Satori} and social networks~\cite{Unicorn,Kineograph}, companies are exploring new knowledge graph architectures. The key advantage of knowledge graphs is that they inherently preserve the structure and semantics of the data. Machine learning on graphs opens new possibilities to discover unknown and unexpected relationships among entities. Ultimately, graph learning systems in engineering attempt to make engineering easy by creating artificial intelligent assistants to co-create with human experts.


Automation engineering data consists of specifications of the product and the production system from different viewpoints such as mechanical, electrical, control, process, manufacturing, defects, and service. Domain-specific tools such as requirements management (RM), computer-aided design (CAD), computer-aided engineering (CAE), and computer-aided manufacturing (CAM) exist to simplify the engineering process and provide interoperability between the different engineering phases and disciplines. Most of these tools organize engineering data in hierarchies such as bill of materials (BOM) and bill of process (BOP), and directed graphs such as requirements, architecture, simulation, control, and software models. A product lifecycle management (PLM) system is typically used to interlink all this data into a large ``graph''. However, there are three main technical problems with existing engineering graphs: (1) sparsity: \textit{within} subgraphs (e.g., a CAD assembly with few parts) and \textit{across} subgraphs (e.g., not all requirements-CAD relationships are available); (2) labeling: although many labels exist (e.g., names, authors, timestamps), these are not consistent and not unique (e.g., two different engineers may name the same thing differently); (3) non-Euclidean: learning on engineering data in the form of graphs, manifolds, and point clouds requires new algorithms and learning architectures suitable for these domains.

\emph{Functional lifting} refers to the process of inferring the functionality of a system from its detailed engineering specifications such as its configuration, code, hybrid equations, geometry, and sensor data. Specifically, a functional model uses high-level modeling abstractions to represent the \textit{purpose} of systems or subsystems~\cite{FMReview}. Today, functional lifting is done manually by engineering experts. This paper introduces a semi-supervised approach to functional lifting by casting the problem to a structure clustering problem in a knowledge graph. Given a set of subgraphs labeled with functional information examples by domain experts, we train a predictor that learns the \textit{structural properties} of the subgraphs and is able to classify these into the known labels. The inference consists of finding the functional score of never seen before subgraphs based on their structural properties. This new capability opens up possibilities for finding functionally equivalent components in product families. It also enables knowledge transfer between different products with a digital thread/trail (e.g., from legacy to under development) under the observation that despite the different components/subsystems, different platforms provide similar functionality.

\begin{figure*}[ht!]
	\centering
    \vspace{-4em}
	\includegraphics[width=0.9\textwidth]{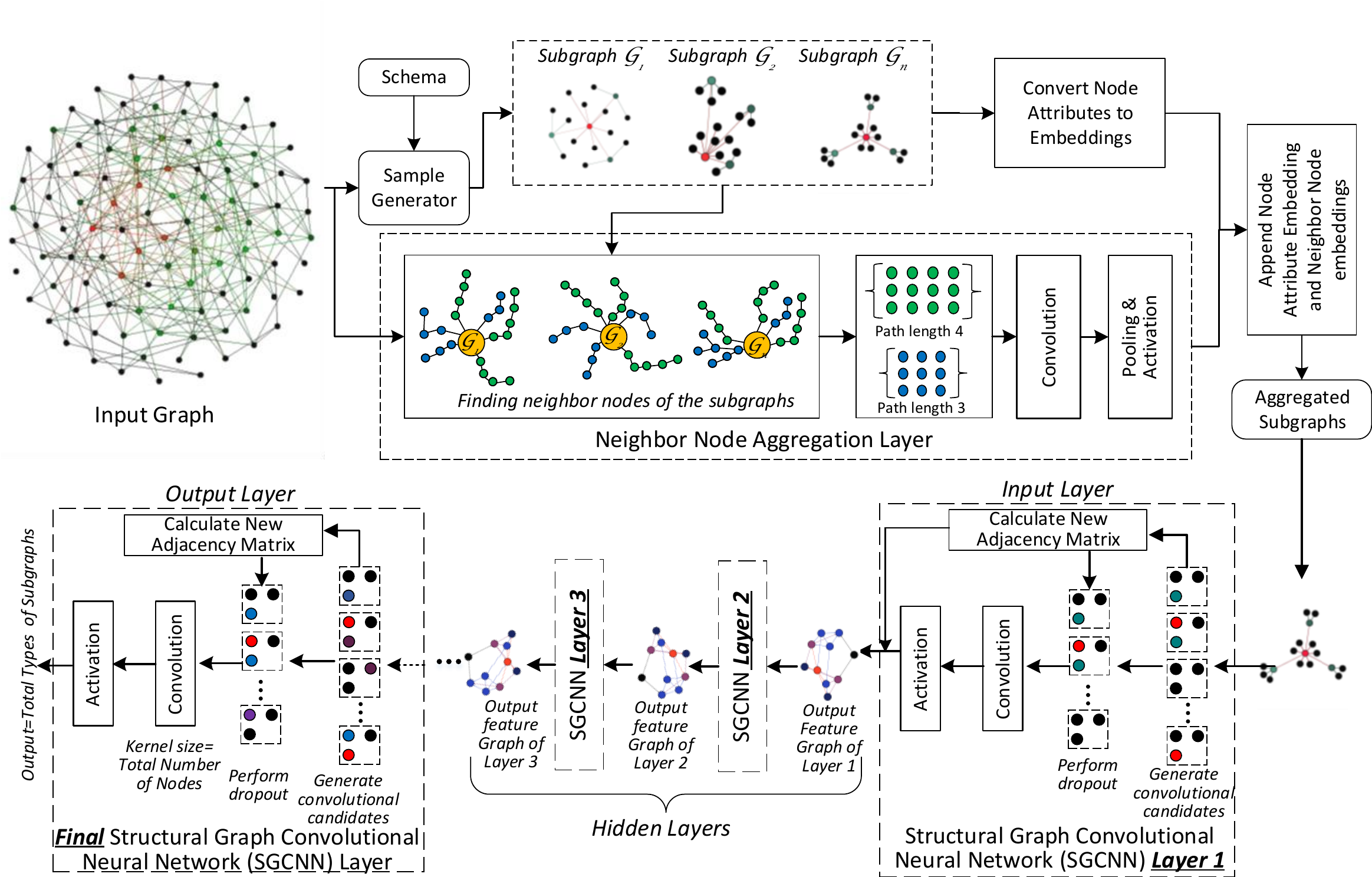}
     \vspace{-0.5em}
	\caption{SGCNN Architecture}
    \vspace{-1em}
	\label{fig:SGCNN_arch}
\end{figure*}
In this paper, we propose a Structured Graph Convolutional Neural Network (SGCNN) that is able to perform graph invariant learning tasks at a graph and subgraph level. The proposed SGCNN is used to automate the functional lifting task from engineering data. The major contributions of SGCNN are as follows:
\begin{itemize}
\item A new path-based neighbor nodes aggregation method that aggregates information from neighbor nodes onto subgraphs for downstream learning.
\item The SGCNN architecture capable of learning at a graph and subgraph level.
\item A novel subgraph convolution kernel for graph invariant convolution operations at a graph and subgraph level.
\item A new downsampling and pooling algorithm based on the local structure of the subgraph.
\end{itemize}

The rest of the paper is organized in six sections. Section~\ref{sec:relatedwork} describes the recent related work in graph learning. Section~\ref{sec:sgcnn} presents the SGCNN architecture and the proposed structural graph learning algorithms. Section~\ref{sec:grabcad} presents the methodology we used to create a representative engineering dataset. Section~\ref{sec:results} evaluates the SGCNN architecture performance. Section~\ref{sec:discussion} discusses potential variations of the SGCNN architecture not covered in this paper. And Section~\ref{sec:conclusions} provides the concluding remarks.

\section{Related work - Graph Learning}\label{sec:relatedwork}
Recently, researchers have made significant breakthroughs in applying convolutional neural networks (CNN) to non-Euclidean structured data such as graphs and manifolds~\cite{tenenbaum2000global, roweis2000nonlinear, belkin2003laplacian, hadsell2006dimensionality, cao2015grarep, henaff2015deep, defferrard2016convolutional, kipf2016semi}. There are two main approaches to CNN on graphs. One is based on spectral domain analysis, the other is based on spatial/vertex domain analysis. In~\cite{defferrard2016convolutional, kipf2016semi}, a Fourier transformation on graphs is proposed to project the high dimension signals, which lives on the vertex of the graph, to low dimension space constructed by the eigenbasis of the graph Laplacian operator~\cite{chung1997spectral}. However, the spectral domain approach has a major limitation: it is not graph invariant. This is because all the spectral domain approaches rely on the Laplacian matrix of the graph. In other words, the Fourier transform on different graph will be different (due to the eigenvectors of Laplacian matrix being graph-dependent), thus a CNN trained for one graph cannot be applied on another one. 

On the other hand, most of the vertex domain approaches are based on the aggregation of the neighborhood information for every node in the graph~\cite{hamilton2017inductive, shervashidze2011weisfeiler, grover2016node2vec}, thus they are graph invariant. For example, in the recent GraphSAGE work~\cite{hamilton2017inductive}, the authors propose to train a function to sample and aggregate a node's local neighborhood to the center node. In this work, the samples are generated in a breadth-first search manner. Several other researchers have followed a similar approach but with distinct sampling methods. For example,~\cite{grover2016node2vec} proposed a factor to tune the sampling ratio between breadth-first search and depth-first search. The vertex domain approach has shown the effectiveness on node-level clustering and classification. However, it also has the limitation that it only works at the node-level. In many real-world engineering problems it is useful to cluster or classify a whole graph or subgraphs instead of a single node.

In addition to CNN on graphs, graph kernels have been used in structure mining to measure the similarity of pairs of graphs~\cite{vishwanathan2010graph}. Although graph kernels can be used to classify or cluster graphs and subgraphs, they only consider the structure similarity between pairs of graphs. In the engineering domain, it is rarely the case that two different structures are represented by the same graph. For example, an electric car and an internal combustion engine car have different drivetrain structures (graphs) but they provide the same function.

\section{Structural Graph Convolutional Neural Networks}\label{sec:sgcnn}
We define a graph as $\mathcal{G} = (\mathcal{V},\mathcal{E})$, where $\mathcal{V}$ is the set of vertices and $\mathcal{E}$ is the set of edges. The graph edges can be weighted and directed. However, for simplicity, this paper only considers unweighted graphs. For each $v_i\in\mathcal{V}$, we define the features to be $f_i$. Features are typically vectors in some higher-dimensional vector space. We define the adjacency matrix of $\mathcal{G}$ to be $\bar{A}$. A subgraph is defined as $\mathcal{G}_s = (\mathcal{V}_s,\mathcal{E}_s)$, where $\mathcal{V}_s \subseteq \mathcal{V}$ and $\mathcal{E}_s\subseteq \mathcal{E}$.

The SGCNN architecture is shown in Figure~\ref{fig:SGCNN_arch}. A \textit{schema} is a user-defined query~\cite{wood2012query} that induces different subgraphs from the input graph. The SGCNN architecture has three major components (highlighted in Figure~\ref{fig:SGCNN_arch}): (1) Neighbor Nodes Aggregation, (2) Subgraph Convolution Kernel, and (3) Graph Pooling. The Neighbor Nodes Aggregation component implements a path-based artificial neural network to aggregate the neighbor nodes information into the target subgraph specified by the schema. The Subgraph Convolution Kernel implements a graph invariant CNN on the target subgraph to extract the subgraph's feature vectors. The Graph Pooling implements a pooling operation to form deep structures in the SGCNN. The details are presented in the following subsections.

\subsection{Attribute Embedding}
The main task of the SGCNN is to learn structure. However, node and edge attributes provide additional information to the learning pipeline. To embed this information in the subgraph we convert node attributes such as descriptions, titles, comments to a vector space using word2vec~\cite{word2vec}. From these embeddings, we generate feature vectors that are used to form the attribute matrix described in Section~\ref{sec:attr_matrix}.

\subsection{Neighbor Nodes Aggregation}
Similar to the vertex domain approach in~\cite{hamilton2017inductive}, given a graph $\mathcal{G}$, the SGCNN aggregates the neighbor nodes features of the target subgraph $\mathcal{G}_t = (\mathcal{V}_t,\mathcal{E}_t)$. Our Neighbor Nodes Aggregation Layer, $l_{ag}$, serves as hidden layers in the SGCNN and uses both the breadth first search and depth first search to collect neighbor nodes in the graph. We define two parameters: $d$ as the depth to search, and $n$ as the number of paths to be computed.

Various other methods for context generation have been proposed. For instance, random walks of a certain length starting at a node provide a notion of context of a node. There is a balance between `local' and `non-local' context generation, commonly referred to as `depth-first' versus `breadth-first.' Neighborhoods provide a breadth-first method of context sampling, while random walks provide more depth. The creators of the node2vec~\cite{grover2016node2vec} node embedding method provide a balanced approach for depth and breadth in context generation. They use a random walk, biased by two parameters $p$ and $q$ where roughly speaking, a large $p$ biases the walk to go `away from home,' while a large $q$ warns the walker of venturing not too far out into the graph. 

\begin{figure}[h]
	\centering
    \vspace{-0.5em}
	\includegraphics[width=0.48\textwidth]{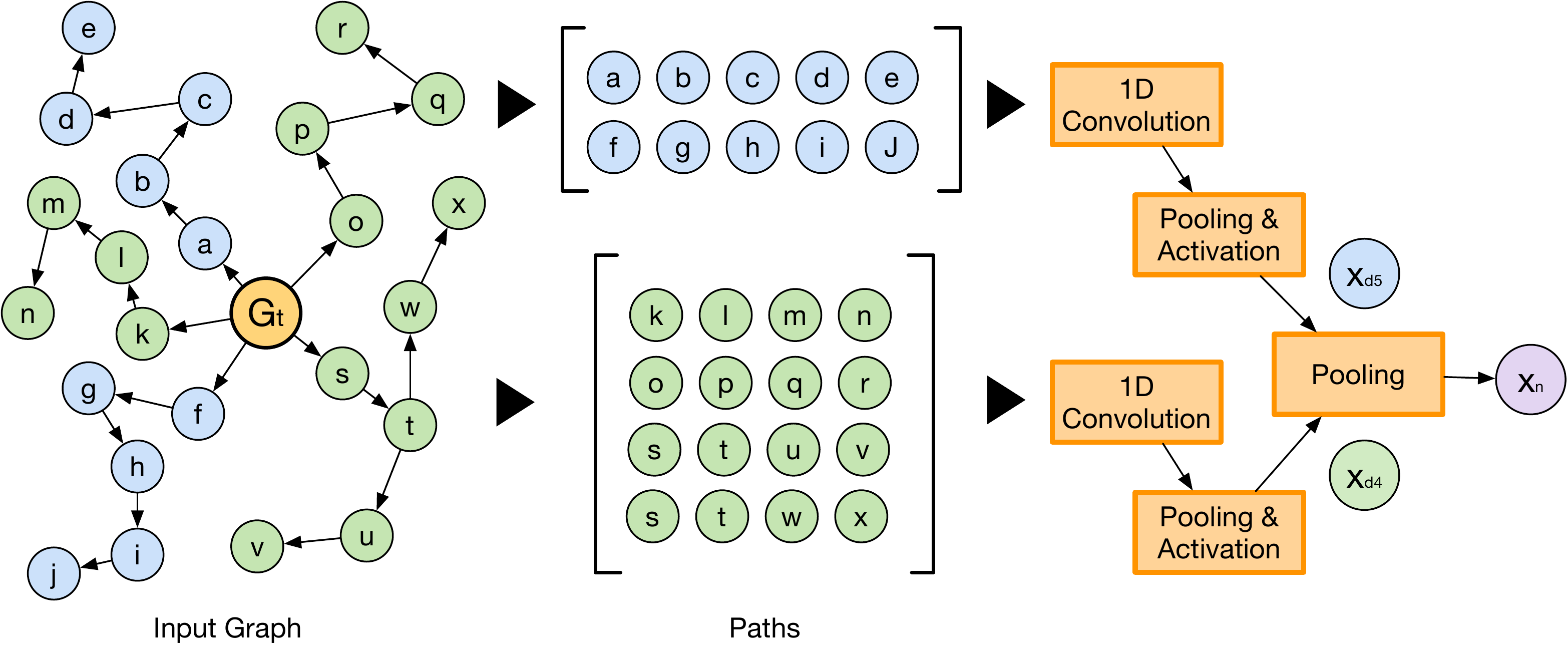}
    \vspace{-1em}
	\caption{Neighbor Nodes Aggregation Example: the yellow node is a simple representation of a subgraph $\mathcal{G}_t$; blue nodes represent length $5$ paths; green nodes represents length $4$ paths.}
	\label{fig:Agg_example}
\end{figure}

For all $v_i\in\mathcal{V}_t$, we search $\mathcal{G}$ to find all length $d$ paths $\mathcal{P}_i^d$, which includes $v_i$, but doesn't include any other nodes in $\mathcal{V}_t$. Then we get all the length $d$ neighbor paths of $\mathcal{G}_t$ as $\mathcal{P}^d = \{\mathcal{P}_0^d\, \mathcal{P}_1^d, ...\}$. From $\mathcal{P}^d$, we randomly select $s$ paths (i.e., we uniformly draw $s$ samples), where $s$ is an input parameter, and use each path as a row to form a neighbor feature matrix $\bar{N}$. Thus $\bar{N}$ is a $n$ by $d$ matrix with each element being a feature vector of a neighbor node. An example of this step is shown in Figure~\ref{fig:Agg_example}. Notice that when the number of paths found in $\mathcal{P}^d$ is smaller than $n$, $\mathcal{P}^d$ can be padded to make $\bar{N}$ with at least $n$ number of rows/paths.

The next task is to extract feature vectors from $\bar{N}$ to generate the output of this layer. As discussed in~\cite{hamilton2017inductive}, the non-Euclidean data such as graphs has no natural ordering. Thus, the feature extraction needs to be applied over an unordered set of paths. In other words, the rows of matrix $\bar{N}$ can be arbitrarily exchanged, and the extracted features should remain unchanged. In this paper, we first apply a general 1-D convolution operation with a trainable $1$ by $d$ weight matrix $\bar{W}$ on $\bar{N}$. As a second step, we use a symmetric pooling function to extract the neighbor nodes feature vectors $x_n$ as follows:

\begin{equation}
x_n = \sigma(f_{pool}(\bar{W} \circledast \bar{N}) + b)
\end{equation}

Where, $b$ is a bias variable, $\sigma$ is an activation function (e.g. ReLU function), and $f_{pool}$ is a pooling function which is invariant to permutations of rows in $\bar{N}$. For example, $f_{pool}$ can be a mean operator over all elements in the matrix, or over all the rows in the matrix. Similarly, $f_{pool}$ can also be a max operator as well. We expect that different $f_{pool}$ functions/operators may be suitable in different domains, and thus it can be a configurable parameter during the training process.

Notice that we can aggregate paths with different lengths. For example, as shown in Figure~\ref{fig:Agg_example}, features from both length $5$ and length $4$ paths are extracted. In general, we extract features from paths with lengths $\{d_1, d_2, d_3, ..., d_k\}$ as $\{x_{d1}, x_{d2}, x_{d3}, ..., x_{dk}\}$. A pooling function $f^d_{pool}$ can also be applied to reduce the dimensions of extracted features as:

\begin{equation}
x_n = f^d_{pool}(\{x_{d1}, x_{d2}, x_{d3}, ..., x_{dk}\})
\end{equation}

Finally, we aggregate $x_n$ to $\mathcal{G}_t$ by concatenating all the feature vectors of $v_i\in\mathcal{V}_t$ as $x_{agg}=\{x_i,x_n\}$. Algorithm~\ref{alg:agg} summarizes our Neighbor Nodes Aggregation process.

\begin{algorithm}[ht]
	\footnotesize
	\SetAlgoLined
	\SetAlgoVlined
	\DontPrintSemicolon
	\KwIn{An input Graph: $G$}
	\KwIn{A schema for query: $Schema$}
	\KwIn{A list of depth to search: $D={d_1, d_2, ..., d_n}$}
	\KwIn{A list of sample numbers per depth: $S={s_1, s_2, ..., s_n}$}
	\KwOut{A feature vector: $x_n$}
	
	Query $G$ with $Schema$ to generate subgraphs $G_t$\;
	
	\ForEach{$d_i\in{D}$}{
		\ForEach{$v_j\in\mathcal{V}_t$}{
			Find all length $d_i$ paths $P^{d_i}_j$\;
			Remove paths containing nodes in $\mathcal{V}_t$ from $P^{d_i}_j$\;
			Add $P^{d_i}_j$ into $P^{d_i}$
		}
		Construct $\bar{N}$ by randomly selecting $s_i$ paths from $P^{d_i}$\;
		Extract feature $x_{di}$ from $\bar{N}$\;
	}
	
	$x_n$ = $f^d_{pool}$($x_{d1},x_{d1},...,x_{dn}$)\;
	
	\Return $x_n$\;
	
	\caption{Neighbor Node Aggregation.}
	\label{alg:agg}
\end{algorithm}

\subsection{SGCNN Layers}
\begin{figure}[h]
	\centering
    \vspace{-1em}
	\includegraphics[width=0.48\textwidth]{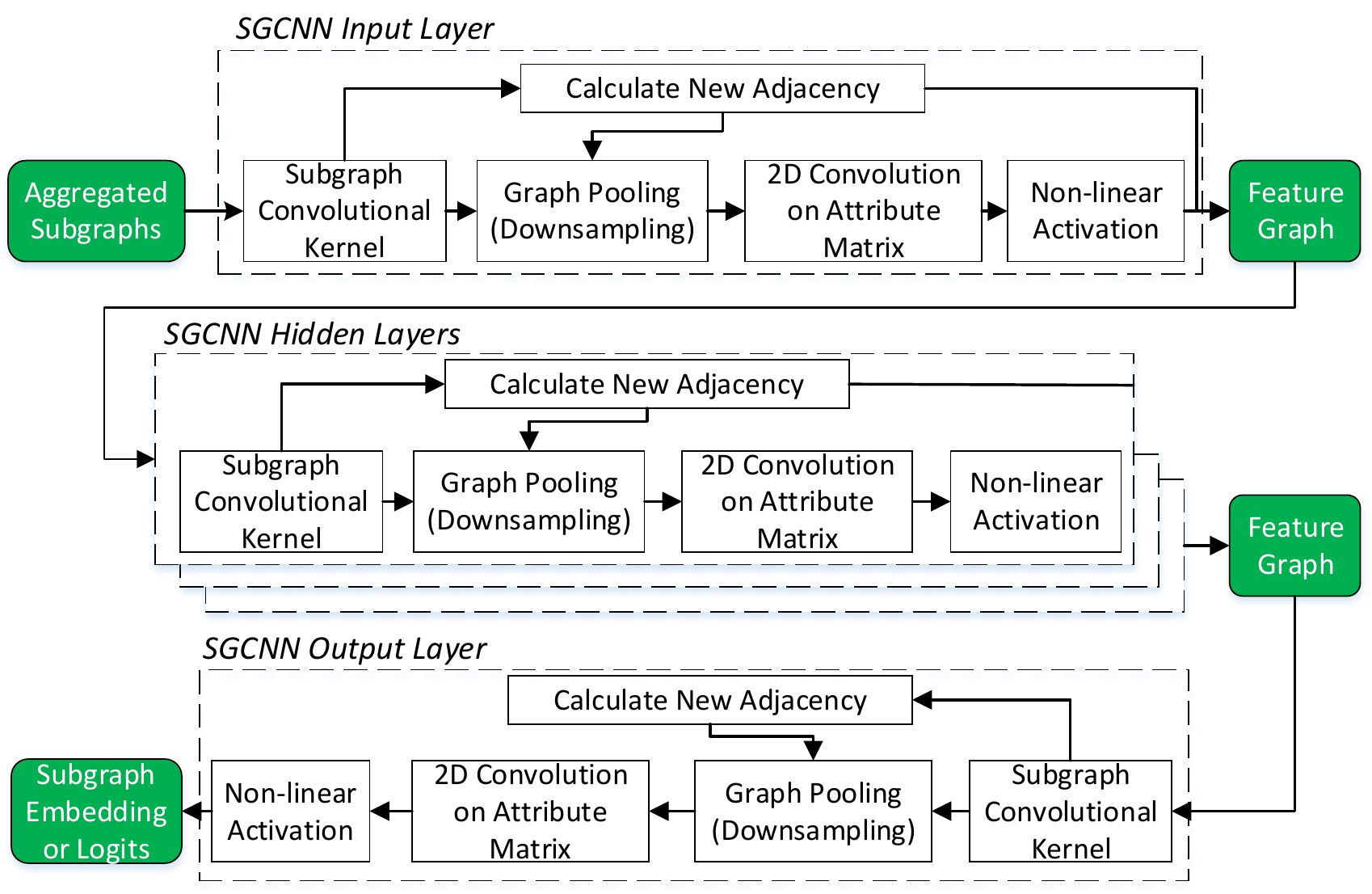}
    \vspace{-1em}
	\caption{Subgraph Convolution Layer Architecture}
	\label{fig:SubConvLayer}
    \vspace{-1em}
\end{figure}

SGCNN can be stacked in layers as shown in Figure~\ref{fig:SubConvLayer}. Each SGCNN layer consists of four subcomponents: (a) Subgraph Convolution Kernel, (b) Graph Pooling, (c) 2D Convolution on Adjacency Matrix, and (d) non-linear activation. The following subsections present each of these subcomponents. 

\subsubsection{Subgraph Convolution Kernel}\label{sec:attr_matrix}
\begin{figure}[h]
	\centering
    \vspace{-1em}
	\includegraphics[width=0.45\textwidth]{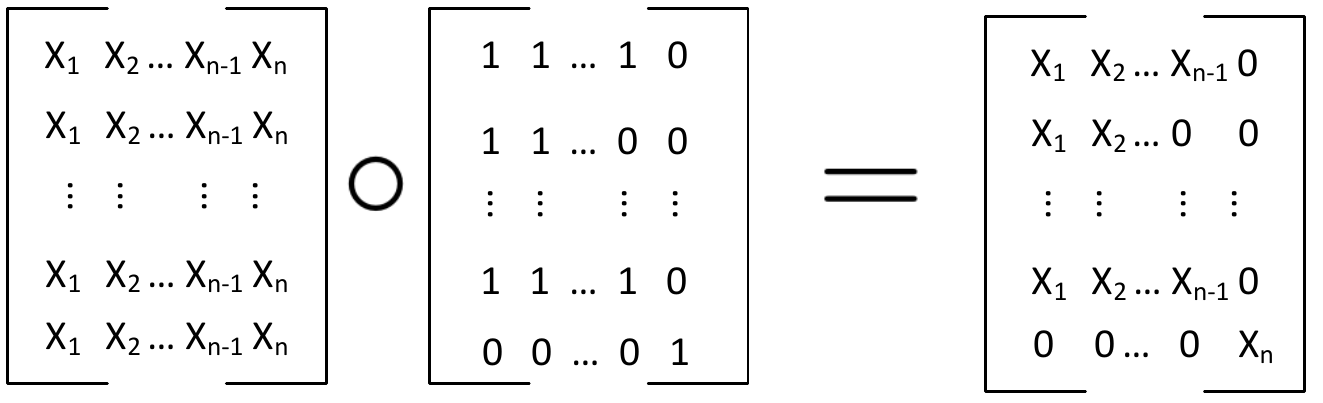}
    \vspace{-1em}
	\caption{Attribute Matrix Example}
	\label{fig:attr_example}
    \vspace{-1em}
\end{figure}

The task of the Subgraph Convolution Kernel is to extract feature vectors from the graph or subgraph. First, we define the attribute matrix of a graph or subgraph as follows. Given a target graph $\mathcal{G}$ with $n$ number of vertices, the list of feature vectors $\mathcal{X}$ is provided by listing -- in any order -- all the feature vectors $x_i$ that exist in $v_i\in\mathcal{V}$. Then, we repeat $\mathcal{X}$ by $n$ times to form the feature matrix $\bar{X}$ with each row being $\mathcal{X}$. With the adjacency matrix $\bar{A}$, we define the attribute matrix $\bar{Ar}$ of $\mathcal{G}$ as the Hadamard Product between $\bar{X}$ and $\bar{A}+\bar{I}$ as follows:

\begin{equation}
\bar{Ar} = \bar{X} \circ (\bar{A}+\bar{I}) 
\end{equation}

where $\bar{I}$ is the identity matrix. For the purpose of maintaining the feature vectors of every vertex without losing their own information~\cite{kipf2016semi}, we add a self-loop to each node by the addition of $\bar{I}$. Notice that although this paper focuses on unweighted graphs, we can easily support weighted graphs by applying a weighted adjacency matrix $\bar{A}$. An example of attribute matrix is shown in Figure~\ref{fig:attr_example}.

Taking $\bar{Ar}$ as the input for the graph convolution operation, we define a graph convolution kernel to be a $k$ by $k$ weight matrix $\bar{W}^k$. Then, we apply the convolution between $\bar{W}^k$ and $\bar{Ar}$. However, different from a grid-structured 2D convolution between matrices (for which the kernel will slide following a top-down, left to right order), we propose a new definition of the convolution operation in a graph data structure.

\begin{algorithm}[h]
	\footnotesize
	\SetAlgoLined
	\SetAlgoVlined
	\DontPrintSemicolon
	\KwIn{An input graph: $G$ with $n$ vertices}
	\KwIn{A convolution kernel: $\bar{W}^k$}
	\KwIn{Sample size: $s$}
	\KwOut{An output feature graph: $G^\prime$}
	
	Generate adjacency matrix $\bar{A}$ from $G$\;
	Using the same vertices order to generate list of features $\mathcal{X}$\;
	Create feature matrix $\bar{X}$ with $n$ rows, and each row being $\mathcal{X}$\;
	
	$\bar{Ar} = \bar{X} \circ (\bar{A}+\bar{I})$\;
	
	$m=\binom{n}{k}$\;
	$CombList$ = Enumerating choice of $n-k$ elements from ${1,2,...,n}$\;
	
	\ForEach{$comb\in{CombList}$}{
		$\bar{Ar}_{comb}$ = remove rows and columns list in $cand$ from $\bar{Ar}$\;
		Add $\bar{Ar}_{comb}$ into $CandList$\;
	}
	
	\If{$m>s$}{
		Down-sample	$CandList$ to $s$ elements\;
	}
	\ElseIf{$m<s$}{
		Pad $CandList$ to $s$ elements\;
	}
	
	\ForEach{$cand\in{CandList}$}{
		$x^k = \bar{W}^k \circledast cand + b$\;
		Add new vertex $v_k$ into $G^\prime$\;
		Add feature vector $x^k$ on $v_k$\;
		Connect $v_k$ based on $cand$'s connection in $G$\;
	}	
	
	\Return $G^\prime$\;
	
	\caption{Graph Convolution Kernel.}
	\label{alg:conv}
\end{algorithm}

Since each row or column in $\bar{Ar}$ is actually corresponding to a vertex in $\mathcal{G}_t$, removing the $i$ row and $i$ column is equivalent to removing the vertex $i$ from $\mathcal{G}_t$. Assuming that $n$ is bigger than $k$, we propose to remove $n-k$ number of vertex from $\mathcal{G}_t$, and the left over subgraph has a new $k$ by $k$ attribute matrix $\bar{Ar}^k$. And there is $\binom{n}{k}$ number of possible $\bar{Ar}^k$. However, the complexity of this operation will be $O\binom{n}{k}$, which is not practical. To relax it, instead of considering all possibilities, we only pick $s$ number of $\bar{Ar}^k$ as a convolution candidate. Thus reducing the complexity to $O(s)$. We achieve this by using a pooling/down-sampling operation (details in Section~\ref{sec:downsample}). 

\subsubsection{Graph Pooling Algorithm}
\label{sec:downsample}
One of the challenges in constructing deep structures for SGCNN is managing the number of kernels in each of the layers. As mentioned earlier, the combinations of all possible convolution candidates is large $O\binom{n}{k}$. Where $n$ is the number of nodes in the subgraph and $k$ is the kernel size. Hence, it becomes computationally infeasible to construct deeper SGCNN layers without performing pooling. To overcome this, we propose the use of a pooling operation before the convolutional operations of each layer. The structure of our proposed subgraph convolution layer is presented in Figure~\ref{fig:SubConvLayer}.

The intuition behind a good down-sampling/pooling operation is to remove the samples with less significant features. Thus, we consider to remove the subgraphs that have lower total degrees compared to other samples for graph structured data. Algorithm~\ref{alg:pooling} describes our proposed graph down-sampling/pooling algorithm.
\vspace{-1em}
\begin{algorithm}[h]
	\footnotesize
	\SetAlgoLined
	\SetAlgoVlined
	\DontPrintSemicolon
	\KwIn{An input graph: $G$}
	\KwIn{The list of the candidate nodes combinations: $Comb$}
	\KwIn{Sample size: $s$}
    \KwIn{Dropout Rate: $d$}
	\KwOut{The list of down-sampled candidate nodes: $Comb^\prime$}
	Randomly sample $Comb^{'}$ from $Comb$ using $d$\;
    Generate adjacency matrix $\bar{A}$ from $G$ for only $Comb^{'}$\;
    \ForEach{$c\in{Comb^{'}}$}{
		$d_c$ = 0\;
		\ForEach{$n\in{c}$}{
			Calculate Degree of $n$ as $d_n$\;
			$d_c=d_c+d_n$\;
		}
		\ForEach{$c^\prime\in{Comb^{'}}$}{
			\If{$c^\prime$ is connected with $c$ in $\bar{A}$}{
				$d_c=d_c+1$\;
			}
		}
	}
	
	Keep the $s$ number of nodes with the highest degrees and store in $Comb^{''}$\;
	
	\Return $Comb^{''}$;
	
	\caption{Graph Pooling Algorithm.}
	\label{alg:pooling}
\end{algorithm}
\vspace{-1em}

Algorithm~\ref{alg:pooling} takes a graph $G$, a list of candidate combinations $Comb$, the pooling sample size $s$, and the dropout rate $d$ as input, and returns the pooled samples of combination list $Comb^{''}$. In Line~1, it randomly samples $Comb^{'}$ by using the dropout rate $d$. This is necessary because it is computationally infeasible to calculate the adjacency matrix in the next step for all the possible candidates in $Comb$. In Line~2, we generate the adjacency matrix $\bar{A}$ for the new candidate combination of nodes. This adjacency matrix carries the graph structural data, and passes it through the deeper layers. This step is important, as the different layers will be able to abstract the graph data in a hierarchical manner. Lines 3-10 compute the total degrees of each candidate nodes combination $Comb^{'}$, which are combinations of the nodes in $G$ that are generated by the convolution kernel and these combinations will serve as the new nodes of the output feature graph after the graph convolution kernel. Specifically, lines 3-7 compute the total degrees inside the combination, and lines 8-10 compute the degrees in between different combinations. Finally, we keep the $s$ number of nodes combinations which have the highest degree, and remove the rest. We significantly reduce the size of the graph convolution kernel by dropping the combinations in the calculated degrees using the max pooling. Nevertheless, we ensure that the convolution is performed on the graph structures with higher connectivity.

\subsubsection{2D Convolutions on Attribute Matrix}
\begin{figure}[h]
	\centering
    \vspace{-1em}
	\includegraphics[width=0.45\textwidth]{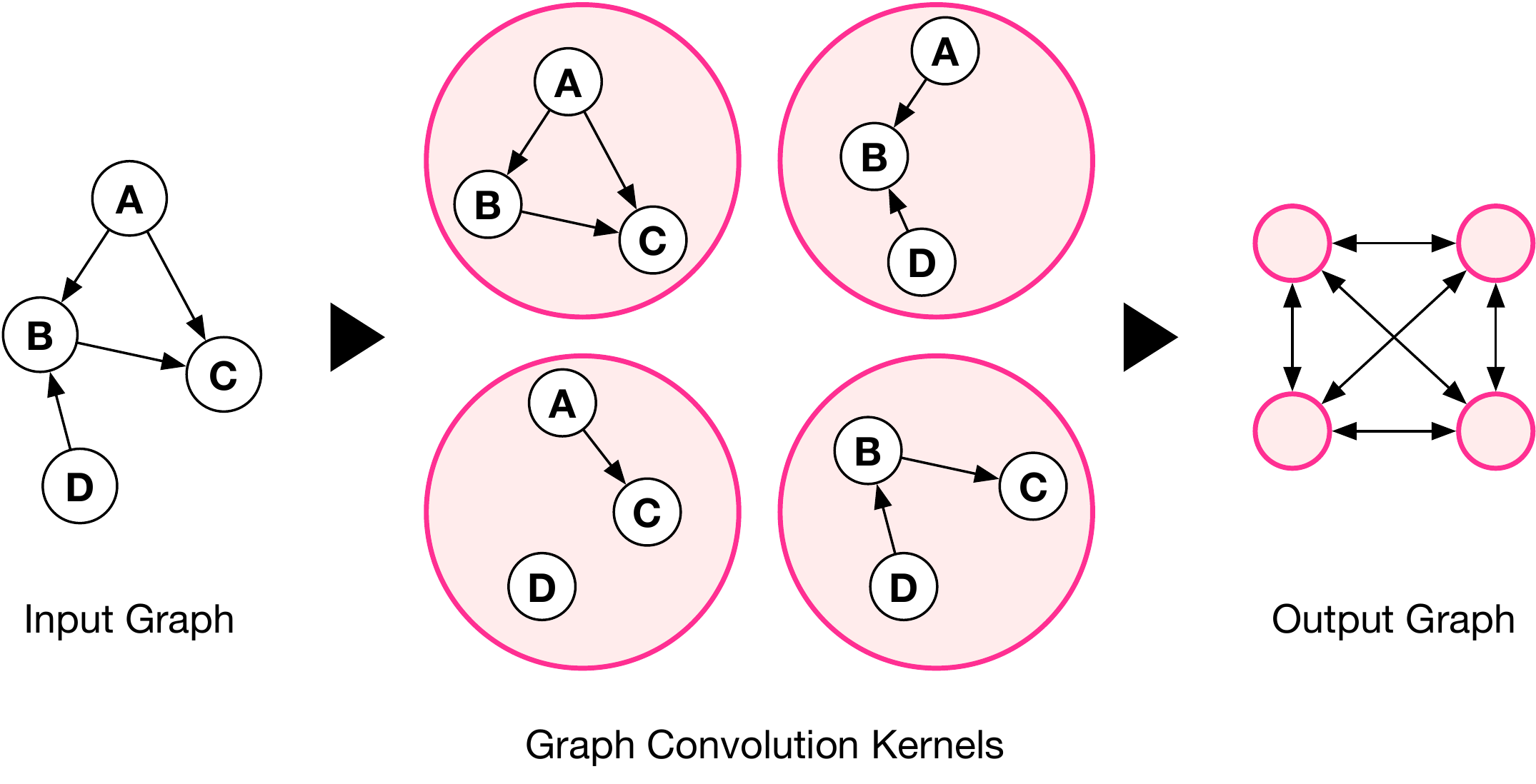}
    \vspace{-1em}
	\caption{Convolution kernel example: given the input graph with 4 vertices, a 3-by-3 convolution kernel is applied. As a result, 4 convolution candidates are generated and the convolution results a new graph with 4 vertices.}
	\label{fig:conv_example}
    \vspace{-1em}
\end{figure}

For all the possible $s$ samples out of $\bar{Ar}^k$, we apply a simple convolution operation to extract feature vectors from this leftover subgraph as follows:

\begin{equation}
x^k = \phi(\bar{W}^k \circledast \bar{Ar}^k + b)
\end{equation}
Where $\phi(.)$ is a non-linear activation function. We will have $s$ extracted feature vectors as: $x^k_1$, $x^k_2$, $...$, $x^k_s$. We consider the extracted feature vectors $x^k_1$, $x^k_2$, $...$, $x^k_s$ as a new feature graph $\mathcal{G}^\prime$ with $s$ number of vertices, and $x^k_i$ as the feature vector for node $i$. An example of this process is shown in Figure~\ref{fig:conv_example}.  

\subsubsection{New Adjacency Matrix Calculation}
$\mathcal{G}^\prime$ can be used as input to another Subgraph Convolution Layer in the proposed SGCNN architecture to form a deep SGCNN model. However, notice that for $\mathcal{G}^\prime$, it is necessary to recalculate the adjacency matrix as well as the attribute matrix. Hence, the new adjacency matrix for each of the  $\mathcal{G}^\prime$ is calculated using the previous adjacency matrix and the corresponding graph convolution kernels. For the new graph, we check the edges between inter and intra nodes of the graph convolution kernels. This new adjacency matrix is then used to perform down sampling and calculate the new attribute matrix. 

\subsection{Classification for Functional Lifting}
Given a large graph and labeled subgraphs, the SGCNN can be used to classify subgraphs. This classification is done by using a softmax function and cross-entropy of the logits. In addition, the feature vectors (subgraph embeddings) generated by the final SGCNN can also be used by clustering algorithms to identify nearest neighbors subgraphs that have an equivalent function in the graph based on their node attributes and structure.  In engineering, there are several use-cases for subgraph embeddings including the identification of functionally equivalent structures that engineers are unaware of, and to identify structures that mislabeled.

\subsection{SGCNN Hyperparameters}
Hyperparameters are the values which are not derived during training of the model but selected prior to training. Hyperparameter selection of deep convolution neural networks for Euclidean data has been extensively studied in~\cite{glorot2010understanding,bergstra2012random}. In addition to the hyperparameters used in deep convolution neural networks such as \textit{activation function, hidden layers, number of iteration, learning rate, and batch size}, the SGCNN has additional hyperparameters that require optimization as well. These hyperparameters are:

\subsubsection{Path length in node aggregation layer:} 
Neighbor node aggregation layer utilizes paths of various lengths to append the subgraph's neighborhood information into the subgraph. Depending on the graph data, the length of the paths can embed different data into the subgraph. Hence, it is necessary to optimize the path length to embed in the subgraph aggregation layer.

\subsubsection{Graph convolution kernel size:} The size of the convolutional kernel not only determines the complexity of the algorithm, but it also determines how the graph features are abstracted. In deep convolutional neural networks, smaller kernel sizes are used to learn local features. Due to the non-eculidean nature of the graph, this paper experiments with various kernel sizes.

\subsubsection{Dropout of candidate kernels:}
In the proposed dropout algorithm, we combine random and adjacency based dropout to tackle the complexity of the learning algorithm. While taking large number of candidate kernels maybe helpful, nodes that are grouped together normally do not have significant connectivity. Hence, the number of candidate kernels is a critical hyperparameter for graph convolutional networks and this paper experiments with different dropout strategies.

\section{GrabCAD Dataset}\label{sec:grabcad}
The vast majority of engineering data is proprietary and therefore not accessible. To evaluate our SGCNN, we generated a dataset from GrabCAD~\footnote{\url{https://grabcad.com/}}. GrabCAD is the largest online community of designers, engineers, and manufacturers where they share 3D CAD models. The GrabCAD community consists of over 4 million members with a library of over 2 million engineering models. We extracted the meta-information from six categories of data consisting of \textit{Car}, \textit{Engine}, \textit{Robotic arm}, \textit{Airplane}, \textit{Gear} and \textit{Wheel} CAD models. From these models we used a schema consisting of the model's name, author, description, parts names, tags, likes, timestamps, and comments and induced subgraphs to generate 2,271 samples for \textit{Car}, 1,597 samples for \textit{Engine}, 2,013 samples for \textit{Robotic arm}, 2,114 samples for \textit{Airplane}, 1,732 samples for \textit{Gear}, and 2,404 samples for \textit{Wheel}. Each of these subgraph consisted of 17 nodes based on the schema with varying number of edges. The subgraphs contained both social network data (e.g. user-to-user relationships through comments and likes) and engineering data (e.g., model-to-tags relationships and model-to-model relationships through common likes and descriptions).

\section{Results}\label{sec:results}
We divided a total of 14,131 samples from different CAD models into 11,304 training samples and 2,827 testing samples. During training we explored various hyperparameters, such as \textit{learning rate}, \textit{batch size}, \textit{kernel size}, \textit{activation function}, \textit{dropout}, etc. The classification accuracy of the six data categories for the SGCNN hyperparameters is show in Table \ref{tab:hyper}. The path length for the node aggregation layer for all the results was set to one. This table shows that the best accuracy is achieved with three layers (node aggregation, SGCNN input layer, and SGCNN output layer) set to a higher learning rate, batch size of 64, large dimension of the hidden layer features, and large output layer kernel size. The final SGCNN layer acts as a dense layer similar to the deep convolutional neural networks.

\begin{table}[h!]
\vspace{-0.5em}
\caption{Accuracy for various hyperparameters.}
\vspace{-1em}
\includegraphics[width=0.45\textwidth]{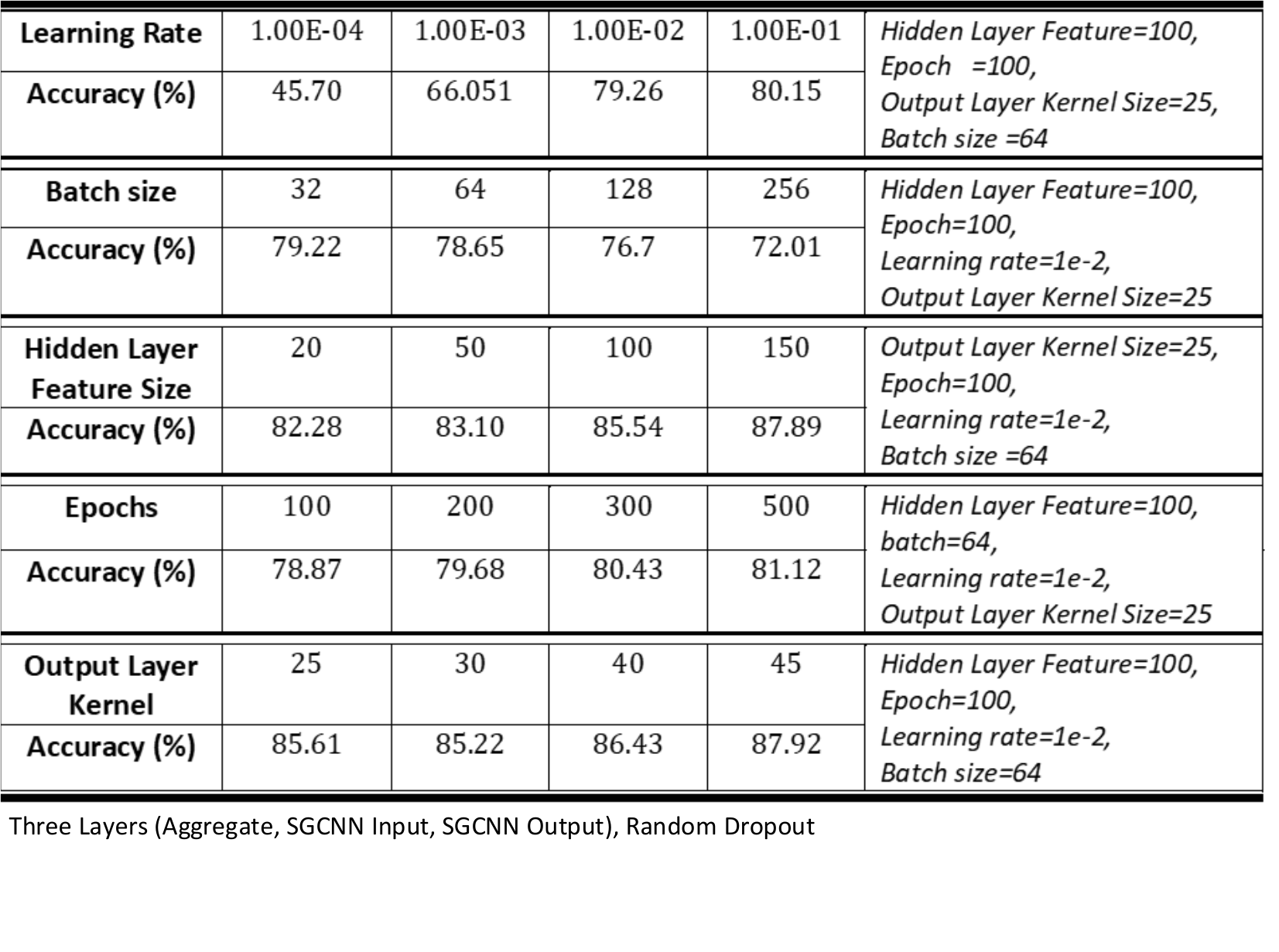}
\label{tab:hyper}
\vspace{-3em}
\end{table}

\subsection{Activation functions}
\begin{figure}[h!]
\vspace{-1em}
\includegraphics[width=0.40\textwidth]{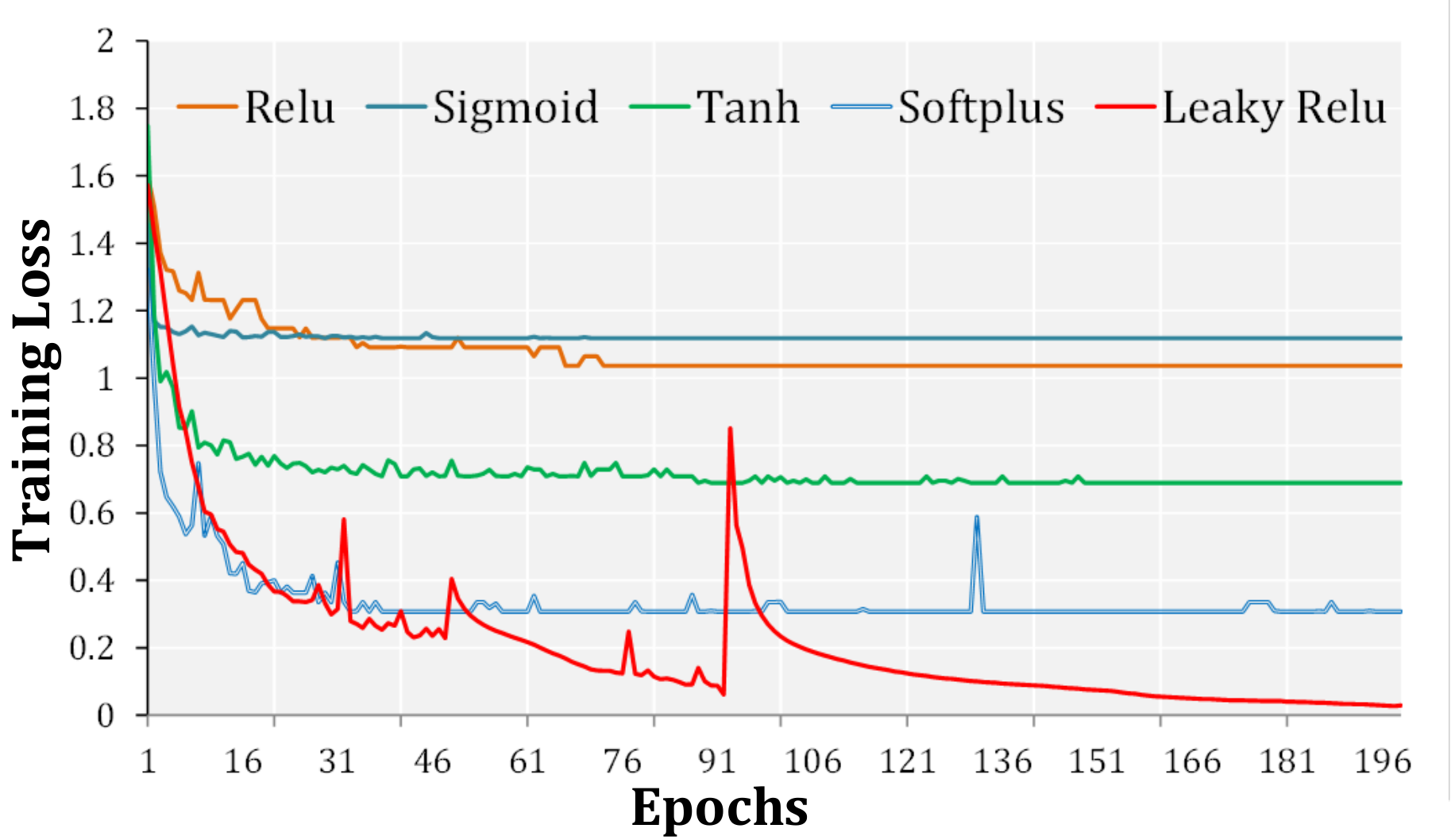}
\vspace{-1em}
\caption{Training loss for different activation functions (layers=2, aggregate and graph embedding layers)}
\vspace{-1em}
\label{fig:actloss}

\end{figure}

\begin{figure}[h!]
\vspace{-1em}
\includegraphics[width=0.40\textwidth]{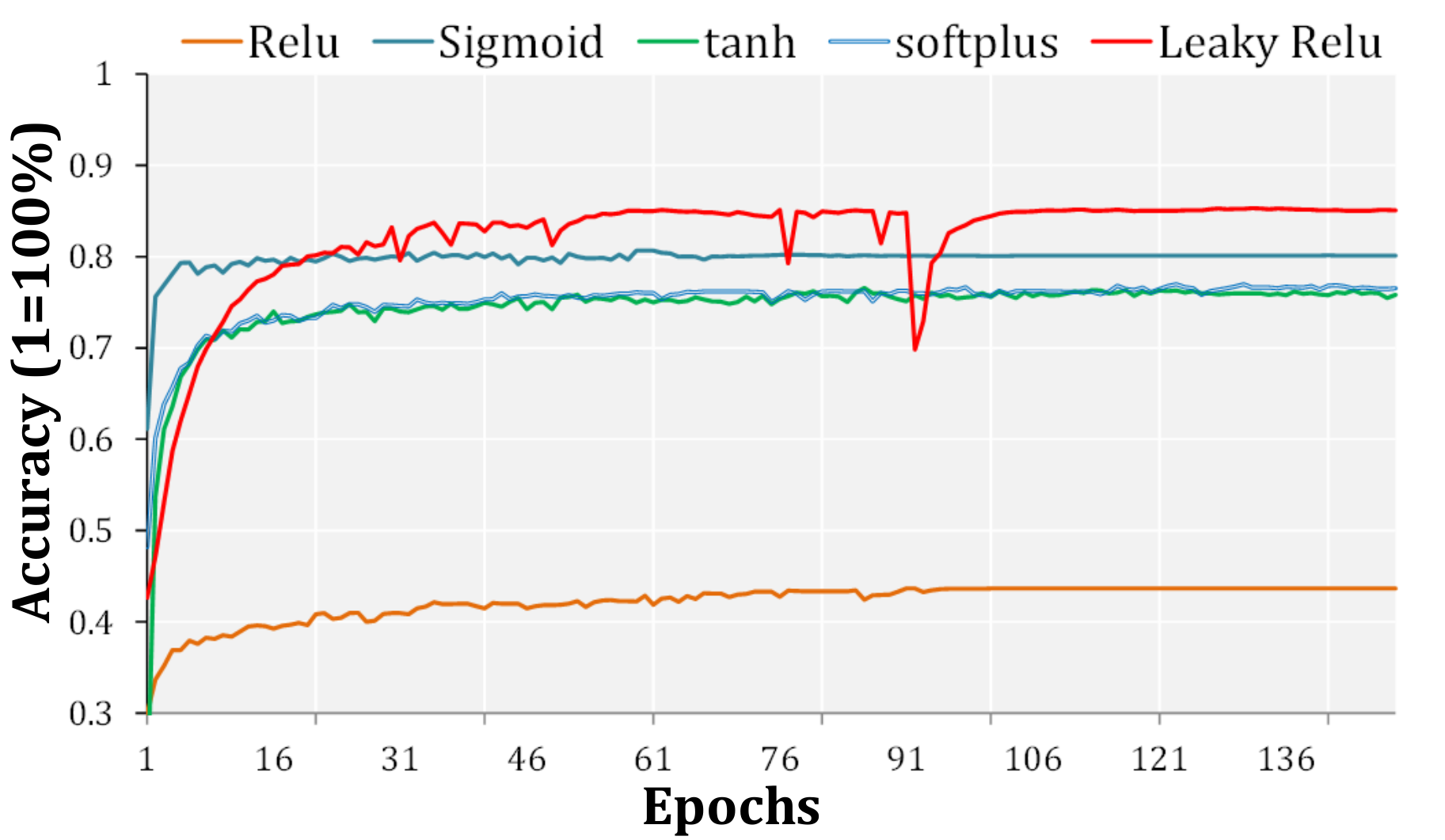}
\vspace{-1em}
\caption{Accuracy loss for different activation functions (layers=2, aggregate and graph embedding layers)}
\label{fig:actacc}
\vspace{-1em}
\end{figure}

The activation functions increase the SGCNN's capacity to learn more complex structures by making it non-linear. We investigated the effects of different activation function (such as \textit{sigmoid}, \textit{softplus}, \textit{tanh}, \textit{rectifier linear unit}, and \textit{leaky rectifier linear unit}) in the testing accuracy (see Fig.~\ref{fig:actacc}) and in training loss (see Fig.~\ref{fig:actloss}).

These results show that the training loss and the subgraph classification accuracy during testing are higher when leaky rectifier linear ($f(x)=\alpha x$ for x<0 and $f(x)=x$ for x>=0) unit is used as an activation function. However, it also introduces some inconsistent loss and accuracy values. The current value of $\alpha$ is 0.2; note that further analysis is required to learn optimized value of $\alpha$.

\subsection{Kernel Size}
\begin{figure}[h]
\vspace{-0.5em}
\includegraphics[width=0.48\textwidth]{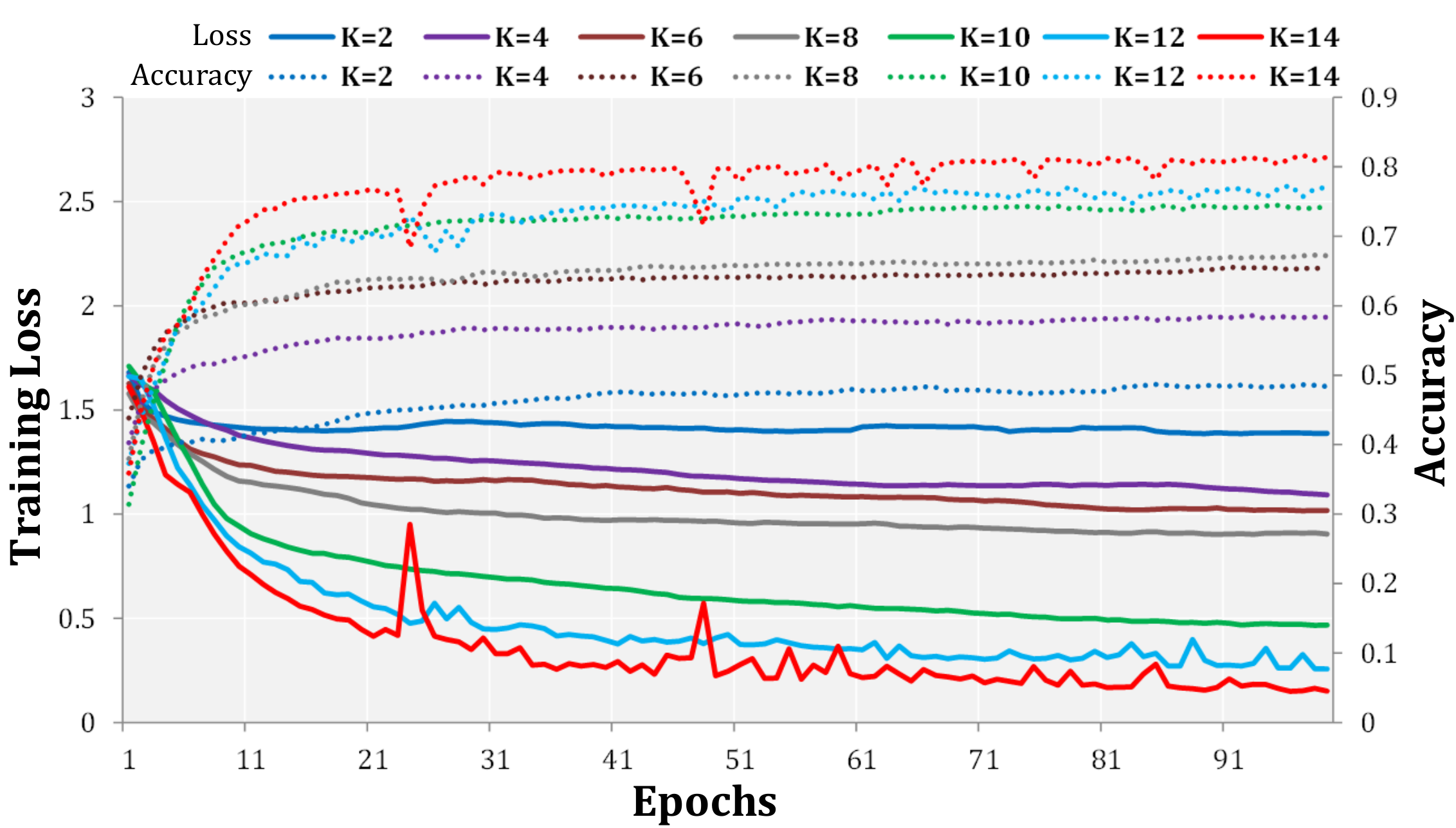}
\vspace{-1em}
\caption{Training loss and accuracy for different size of kernels (with random dropout, last layer kernel size=5, layers=3, hidden layer activation= relu, final layer activation=leaky relu).}
\label{fig:grabcad}
\vspace{-1em}
\end{figure}

The kernel size determines the number nodes considered in the graph convolution. Figure~\ref{fig:grabcad} shows the effects of using various kernel sizes in a shallow, three layer SGCNN consisting of node-aggregation layer, SGCNN input layer and an SGCNN output layer. These results show that the kernel size (2, 4, 6, 8, 10, 12, 14) has a significant effect in the performance of the model. A large kernel size achieves higher classification accuracy and lower training loss. 

\subsection{Dropout}
\begin{figure}[h]
\vspace{-1em}
\includegraphics[width=0.48\textwidth]{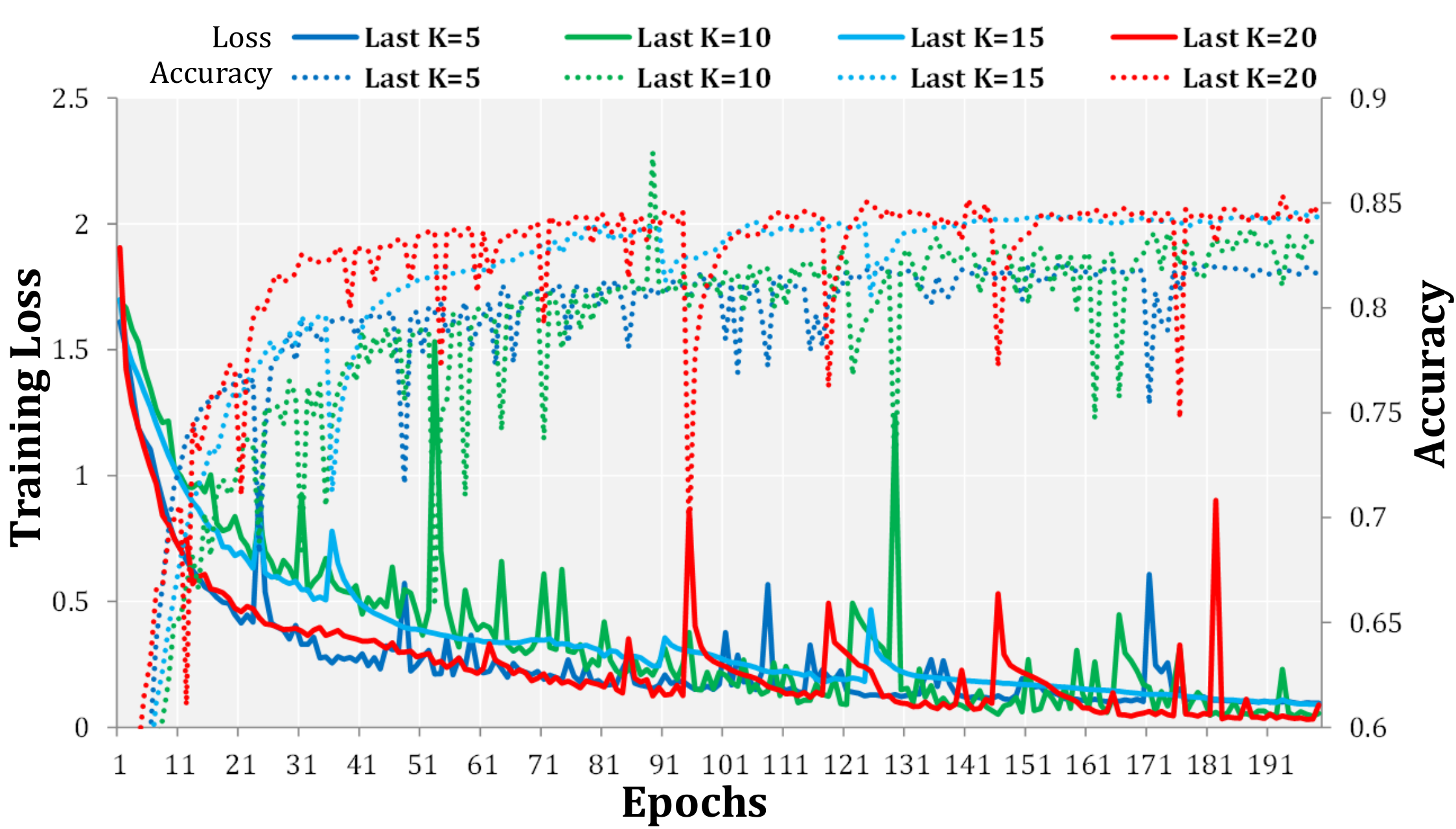}
\vspace{-1em}
\caption{Training loss and accuracy for various random dropouts to match the last layer's kernel size (hidden layer kernel size=14, layers=3, hidden layer activation=relu, final layer activation=leaky relu)}
\label{fig:lastk}
\vspace{-1em}
\end{figure}

Dropout is essential for the scalability of the SGCNN into deeper layers with different kernel and subgraph sizes. In the proposed methodology, we lower the timing complexity by using the graph pooling algorithm (Section~\ref{sec:downsample}) that also performs random dropouts before calculating the new adjacency matrix. Fig.~\ref{fig:lastk} shows the results for different random dropouts.

The dropout is based on the kernel size of the SGCNN output layer. The selected subgraph node size is 17, and the kernel size of SGCNN input layer is 14, which means the possible candidate for the number of kernel is $\binom{17}{14}=680$. We demonstrate the result of performing random dropout on three layers of SGCNN. We have dropped the total graph convolutional kernel to 5, 10, 15, and 20 by setting the kernel size of the SGCNN output layer to the respective values. Fig.~\ref{fig:lastk} shows that the training loss decreases and the testing classification accuracy increases with lower dropout values. Although the dropout rate is quite high, the kernel size of 14 in the SGCNN input layer already encompasses most of the nodes in subgraph, allowing the model to learn the classes of the subgraphs.

\begin{figure}[h!]
\vspace{-1em}
\includegraphics[width=0.48\textwidth]{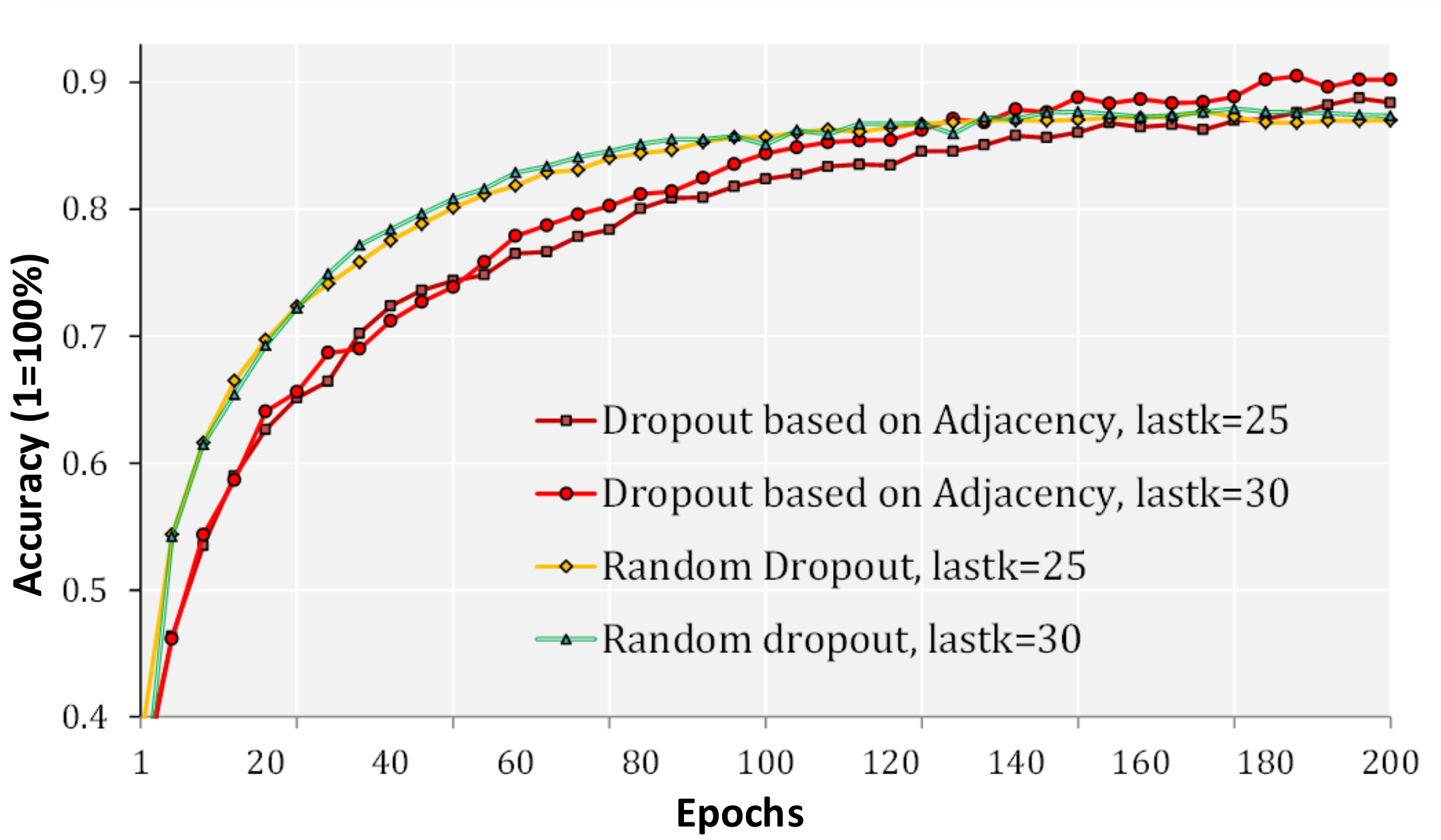}
\vspace{-1em}
\caption{Comparison between random dropout and adjacency based dropout (hidden layer kernel size=14, layers=3, hidden layer activation=relu, final layer activation=leaky relu).}
\vspace{-1em}
\label{fig:dropout}
\end{figure}
Fig.~\ref{fig:dropout} shows a comparison of random dropout versus the proposed down sampling algorithm. The down sampling algorithm initially performs random dropout of the candidate graph convolutional kernels to 50. Then, the new adjacency is calculated and the candidate kernels are dropped further to 25 and 35. It may be noticed that although random dropout is able to achieve higher accuracy during initial epochs, the proposed down-sampling algorithm achieves accuracy of more than 90\%. Furthermore increasing the initial random dropout may improve the accuracy but at the cost of increasing the timing complexity for calculation of bigger adjacency matrix of the new graph.

\begin{figure}[h!]
\vspace{-1em}
\includegraphics[width=0.48\textwidth]{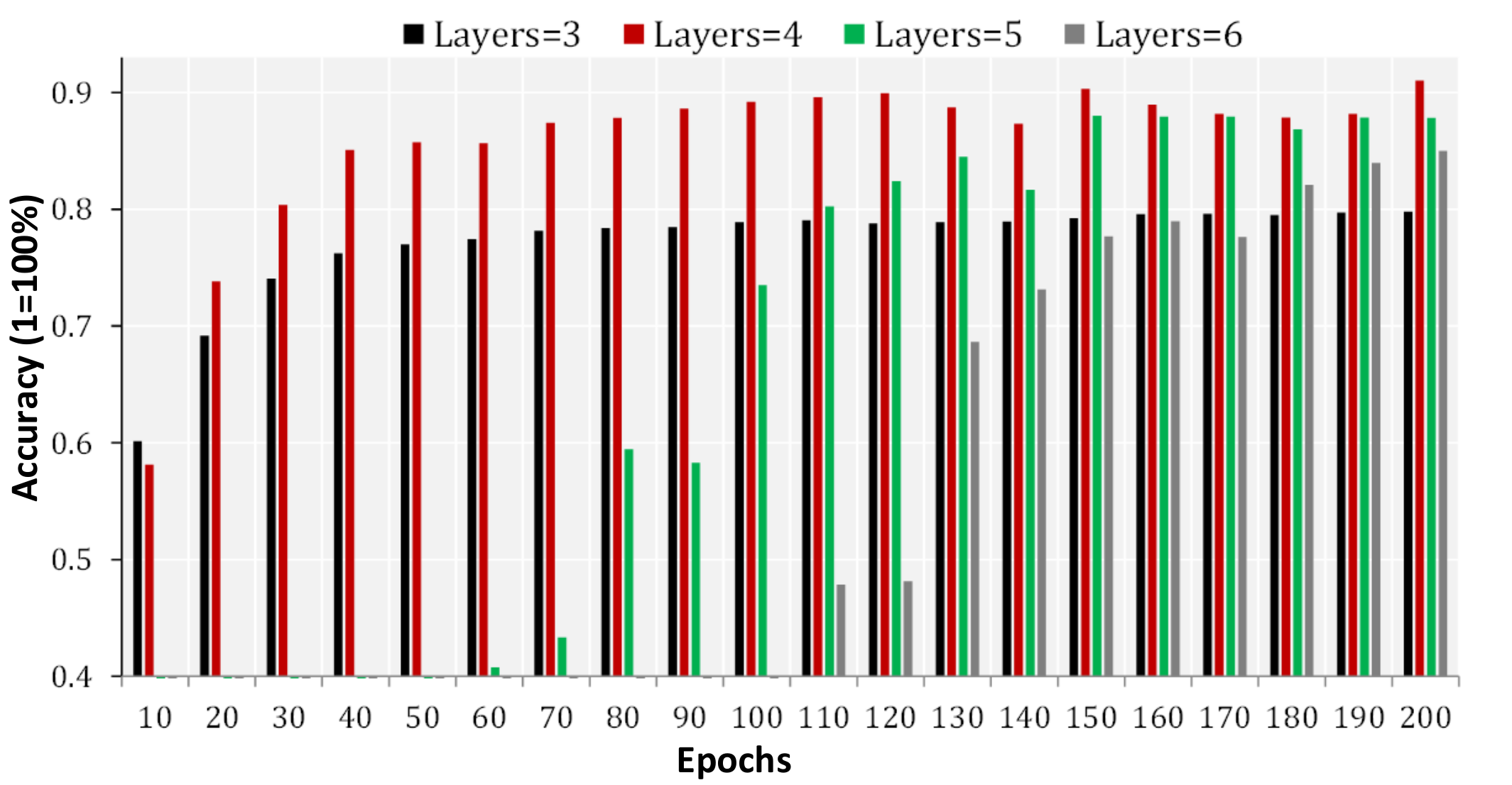}
\vspace{-1em}
\caption{Performance comparison between various layer sizes (hidden layer kernel size=2, activation =leaky relu, lastk=30, Dropout=Random)}
\vspace{-1em}
\label{fig:layers}
\end{figure}

\subsection{Layers}
In Figure \ref{fig:layers}, we explore deeper SGCNN layers with additional hidden layers. With three layers, the model consists of only node aggregation layer, SGCNN input layer, and a SGCNN output layer. For additional hidden, deeper layers, the kernel size of 2 is selected. In each of the consecutive layers, the model learns the structural features in an hierarchical manner by merging information from smaller node sizes. Fig.~\ref{fig:layers} shows that compared to shallow three layers, deeper layers achieve higher accuracy. The highest classification accuracy is $\approx$91\% with four layers.

\section{Discussion}\label{sec:discussion}
The SGCNN's capability to learn subgraph structure embedded with attributes was demonstrated in Section~\ref{sec:results}. It achieves very positive results on functional lifting using the GrabCAD dataset. However, there are some aspects that were not covered in the paper and we briefly discuss in this section.

\subsection{Possible variations of SGCNN}
The presented SGCNN architecture consists of the node aggregation layer followed by the SGCNN input layer. A possible variation is to have a node aggregation layer after the SGCNN layers. This variation would possibly converge faster.
Currently, the schema is provided by the user and thus requires domain expertise. A possible variation is to add the schema generation in the SGCNN learning pipeline using unsupervised learning methods. This variation would provide a more general approach for functional lifting.

\subsection{Hyperparameter Optimization}
To validate our initial hypothesis in structural subgraph learning, we presented several results for hyperparameter variations. However, exhaustive exploration of the hyperparameter is subject to the continuation of this work. The most important hyperparameters are path length in the node aggregation layer, dropout based on adjacency based on different layer sizes and initial random dropout, and deep SGCNN layers with different convolution kernel sizes space. Furthermore, in our future work, we will validate the SGCNN hyperparameters on different graph data sets.

\subsection{Comparison with related work}
Although SGCNN was created to address the functional lifting problem in engineering, we believe this is broadly applicable to other domains. In our future work, we will compare the performance of SGCNN against the latest work on graph convolutional networks targeting subgraph-level embeddings \cite{narayanan2016subgraph2vec,adhikari2018sub2vec}.

\section{Conclusions}\label{sec:conclusions}
This paper proposes a novel graph learning method for engineering data. The SGCNN architecture was developed to perform graph invariant learning tasks at a graph and subgraph levels. Using a realistic engineering-related dataset consisting of 14,131 graph samples, the SGCNN is capable of performing subgraph classification of functionality with an accuracy of 91\%. The key insights are: (a) for shallow SGCNN networks, large kernel sizes are necessary for high-accuracy; (b) for deep SGCNN networks, small kernel sizes are capable of learning functional abstractions in a hierarchical manner; and (c) even with more than 50\% dropout, SGCNN is able to learn the structure of the subgraphs.

\bibliographystyle{abbrv}
\bibliography{biblio}

\begin{thebibliography}{10}

\bibitem{adhikari2018sub2vec}
B.~Adhikari, Y.~Zhang, N.~Ramakrishnan, and B.~A. Prakash.
\newblock Sub2vec: Feature learning for subgraphs.
\newblock In {\em Pacific-Asia Conference on Knowledge Discovery and Data
  Mining}, pages 170--182. Springer, 2018.

\bibitem{belkin2003laplacian}
M.~Belkin and P.~Niyogi.
\newblock Laplacian eigenmaps for dimensionality reduction and data
  representation.
\newblock {\em Neural computation}, 15(6):1373--1396, 2003.

\bibitem{bergstra2012random}
J.~Bergstra and Y.~Bengio.
\newblock Random search for hyper-parameter optimization.
\newblock {\em Journal of Machine Learning Research}, 13(Feb):281--305, 2012.

\bibitem{cao2015grarep}
S.~Cao, W.~Lu, and Q.~Xu.
\newblock Grarep: Learning graph representations with global structural
  information.
\newblock In {\em Proceedings of the 24th ACM International on Conference on
  Information and Knowledge Management}, pages 891--900. ACM, 2015.

\bibitem{Kineograph}
R.~Cheng, J.~Hong, A.~Kyrola, Y.~Miao, X.~Weng, M.~Wu, F.~Yang, L.~Zhou,
  F.~Zhao, and E.~Chen.
\newblock Kineograph: Taking the pulse of a fast-changing and connected world.
\newblock In {\em Proceedings of the 7th ACM European Conference on Computer
  Systems}, EuroSys '12, pages 85--98, New York, NY, USA, 2012. ACM.

\bibitem{chung1997spectral}
F.~R. Chung.
\newblock {\em Spectral graph theory}.
\newblock Number~92. American Mathematical Soc., 1997.

\bibitem{Unicorn}
M.~Curtiss, I.~Becker, T.~Bosman, S.~Doroshenko, L.~Grijincu, T.~Jackson,
  S.~Kunnatur, S.~Lassen, P.~Pronin, S.~Sankar, G.~Shen, G.~Woss, C.~Yang, and
  N.~Zhang.
\newblock Unicorn: A system for searching the social graph.
\newblock {\em Proc. VLDB Endow.}, 6(11):1150--1161, Aug. 2013.

\bibitem{defferrard2016convolutional}
M.~Defferrard, X.~Bresson, and P.~Vandergheynst.
\newblock Convolutional neural networks on graphs with fast localized spectral
  filtering.
\newblock In {\em Advances in Neural Information Processing Systems}, pages
  3844--3852, 2016.

\bibitem{KnowledgeVault}
X.~Dong, E.~Gabrilovich, G.~Heitz, W.~Horn, N.~Lao, K.~Murphy, T.~Strohmann,
  S.~Sun, and W.~Zhang.
\newblock Knowledge vault: A web-scale approach to probabilistic knowledge
  fusion.
\newblock In {\em Proceedings of the 20th ACM SIGKDD International Conference
  on Knowledge Discovery and Data Mining}, KDD '14, pages 601--610, New York,
  NY, USA, 2014. ACM.

\bibitem{FMReview}
M.~Erden, H.~Komoto, T.~V. Beek, V.D'Amelio, E.~Echavarria, and T.~Tomiyama.
\newblock A review of function modeling: approaches and applications.
\newblock {\em Artificial Intelligence for Engineering Design, Analysis and
  Manufacturing}, 22:147--169, 2008.

\bibitem{glorot2010understanding}
X.~Glorot and Y.~Bengio.
\newblock Understanding the difficulty of training deep feedforward neural
  networks.
\newblock In {\em Proceedings of the thirteenth international conference on
  artificial intelligence and statistics}, pages 249--256, 2010.

\bibitem{grover2016node2vec}
A.~Grover and J.~Leskovec.
\newblock node2vec: Scalable feature learning for networks.
\newblock In {\em Proceedings of the 22nd ACM SIGKDD international conference
  on Knowledge discovery and data mining}, pages 855--864. ACM, 2016.

\bibitem{hadsell2006dimensionality}
R.~Hadsell, S.~Chopra, and Y.~LeCun.
\newblock Dimensionality reduction by learning an invariant mapping.
\newblock In {\em Computer vision and pattern recognition, 2006 IEEE computer
  society conference on}, volume~2, pages 1735--1742. IEEE, 2006.

\bibitem{hamilton2017inductive}
W.~Hamilton, Z.~Ying, and J.~Leskovec.
\newblock Inductive representation learning on large graphs.
\newblock In {\em Advances in Neural Information Processing Systems}, pages
  1025--1035, 2017.

\bibitem{henaff2015deep}
M.~Henaff, J.~Bruna, and Y.~LeCun.
\newblock Deep convolutional networks on graph-structured data.
\newblock {\em arXiv preprint arXiv:1506.05163}, 2015.

\bibitem{kipf2016semi}
T.~N. Kipf and M.~Welling.
\newblock Semi-supervised classification with graph convolutional networks.
\newblock {\em arXiv preprint arXiv:1609.02907}, 2016.

\bibitem{word2vec}
T.~Mikolov, I.~Sutskever, K.~Chen, G.~Corrado, and J.~Dean.
\newblock Distributed representations of words and phrases and their
  compositionality.
\newblock In {\em Proceedings of the 26th International Conference on Neural
  Information Processing Systems - Volume 2}, NIPS'13, pages 3111--3119, USA,
  2013. Curran Associates Inc.

\bibitem{narayanan2016subgraph2vec}
A.~Narayanan, M.~Chandramohan, L.~Chen, Y.~Liu, and S.~Saminathan.
\newblock subgraph2vec: Learning distributed representations of rooted
  sub-graphs from large graphs.
\newblock {\em arXiv preprint arXiv:1606.08928}, 2016.

\bibitem{roweis2000nonlinear}
S.~T. Roweis and L.~K. Saul.
\newblock Nonlinear dimensionality reduction by locally linear embedding.
\newblock {\em science}, 290(5500):2323--2326, 2000.

\bibitem{shervashidze2011weisfeiler}
N.~Shervashidze, P.~Schweitzer, E.~J.~v. Leeuwen, K.~Mehlhorn, and K.~M.
  Borgwardt.
\newblock Weisfeiler-lehman graph kernels.
\newblock {\em Journal of Machine Learning Research}, 12(Sep):2539--2561, 2011.

\bibitem{tenenbaum2000global}
J.~B. Tenenbaum, V.~De~Silva, and J.~C. Langford.
\newblock A global geometric framework for nonlinear dimensionality reduction.
\newblock {\em science}, 290(5500):2319--2323, 2000.

\bibitem{vishwanathan2010graph}
S.~V.~N. Vishwanathan, N.~N. Schraudolph, R.~Kondor, and K.~M. Borgwardt.
\newblock Graph kernels.
\newblock {\em Journal of Machine Learning Research}, 11(Apr):1201--1242, 2010.

\bibitem{wood2012query}
P.~T. Wood.
\newblock Query languages for graph databases.
\newblock {\em ACM SIGMOD Record}, 41(1):50--60, 2012.

\bibitem{Satori}
F.~Zhang, N.~J. Yuan, D.~Lian, X.~Xie, and W.-Y. Ma.
\newblock Collaborative knowledge base embedding for recommender systems.
\newblock In {\em Proceedings of the 22Nd ACM SIGKDD International Conference
  on Knowledge Discovery and Data Mining}, KDD '16, pages 353--362, New York,
  NY, USA, 2016. ACM.

\end{thebibliography}

\end{document}